\documentclass{article}

\PassOptionsToPackage{numbers, compress}{natbib}

 \usepackage[preprint]{neurips_2025}

\usepackage[utf8]{inputenc} % allow utf-8 input
\usepackage[T1]{fontenc}    % use 8-bit T1 fonts
\usepackage{hyperref}       % hyperlinks
\usepackage{url}            % simple URL typesetting
\usepackage{booktabs}       % professional-quality tables
\usepackage{amsfonts}       % blackboard math symbols
\usepackage{nicefrac}       % compact symbols for 1/2, etc.
\usepackage{microtype}      % microtypography
\usepackage{xcolor}         % colors
\usepackage{graphicx}
\usepackage{subcaption}
\usepackage{floatrow}
\usepackage{multirow}
\usepackage{booktabs}  % 提供\toprule等命令
\usepackage{caption}   % 自定义标题样式（可选）
\usepackage{wrapfig}
\newfloatcommand{capbtabbox}{table}[][\FBwidth]

\title{CrowdTrack: A Benchmark for Difficult Multiple Pedestrian Tracking in Real Scenarios}

\author{%
  Teng Fu\\
  Fudan University\\
  \texttt{tfu23@m.fudan.edu.cn} \\
  \And
  Yuwen Chen \\
  Fudan University \\
  \texttt{ywchen23@m.fudan.edu.cn} \\
  \AND
  Zhuofan Chen \\
  Fudan University \\
  \texttt{zfchen23@m.fudan.edu.cn} \\
  \And
  Mengyang Zhao \\
  Fudan University \\
  \texttt{myzhao@fudan.edu.cn} \\
  \And
  Bin Li\thanks{Corresponding authors} \\
  Fudan University \\
  \texttt{libin@fudan.edu.cn} \\
  \And
  Xiangyang Xue\footnotemark \\
  Fudan University \\
  \texttt{xyxue@fudan.edu.cn} \\
}

\begin{document}

\let\oldthanks\thanks
\renewcommand{\thanks}[1]{\oldthanks{#1}\setcounter{footnote}{0}}
\maketitle
% \corauthnote
% \renewcommand{\thefootnote}{}
% \footnotetext{ $^{*}$ ~Corresponding authors.}

\begin{abstract}
    Multi-object tracking is a classic field in computer vision. Among them, pedestrian tracking has extremely high application value and has become the most popular research category. Existing methods mainly use motion or appearance information for tracking, which is often difficult in complex scenarios. For the motion information, mutual occlusions between objects often prevent updating of the motion state; for the appearance information, non-robust results are often obtained due to reasons such as only partial visibility of the object or blurred images. Although learning how to perform tracking in these situations from the annotated data is the simplest solution, the existing MOT dataset fails to satisfy this solution. Existing methods mainly have two drawbacks: relatively simple scene composition and non-realistic scenarios. Although some of the video sequences in existing dataset do not have the above-mentioned drawbacks, the number is far from adequate for research purposes. To this end, we propose a difficult large-scale dataset for multi-pedestrian tracking, shot mainly from the first-person view and all from real-life complex scenarios. We name it ``CrowdTrack'' because there are numerous objects in most of the sequences. Our dataset consists of 33 videos, containing a total of 5,185 trajectories. Each object is annotated with a complete bounding box and a unique object ID. The dataset will provide a platform to facilitate the development of algorithms that remain effective in complex situations. We analyzed the dataset comprehensively and tested multiple SOTA models on our dataset. Besides, we analyzed the performance of the foundation models on our dataset. The dataset and project code is released at: \url{https://github.com/loseevaya/CrowdTrack}
\end{abstract}

\section{Introduction}
Multi-Object Tracking (MOT)~\cite{bewley2016simple, wojke2017simple, fu2025foundation, cai2022memot, meinhardt2022trackformer} remains a fundamental challenge in computer vision, requiring the prediction of object trajectories in continuous image sequences while preserving consistent identity labels across frames. Among diverse tracking targets, pedestrian tracking has garnered substantial research attention due to its critical applications in embodied intelligence, autonomous driving, and video surveillance. Existing MOT approaches predominantly follow two paradigms: $(1)$ The tracking-by-detection (TBD) paradigm~\cite{zhang2022bytetrack, zhang2021fairmot, cao2023observation, aharon2022bot, yang2023hard}, which relies on pretrained object detectors to generate bounding-box predictions and subsequently links detections across frames via location or appearance-based association strategies; $(2)$ The end-to-end Transformer-based paradigm~\cite{fu2023denoising, cai2022memot, meinhardt2022trackformer, sun2020transtrack}, which maintains tracklet-specific hidden states as special queries and directly regresses object locations by fusing cross-frame features through attention mechanisms.
The TBD pipeline typically decouples detection and tracking, leveraging mature detection backbones (e.g., YOLOX~\cite{ge2021yolox}, Faster R-CNN~\cite{ren2015faster}) but suffering from error accumulation due to sequential dependency on detection outputs. In contrast, end-to-end models like TrackFormer~\cite{meinhardt2022trackformer} encode temporal context directly via Transformer layers, enabling joint learning of object localization and trajectory association. However, both paradigms face challenges in handling occlusions, scale variations, and low-resolution scenarios—common in real-world pedestrian tracking datasets.
% Multi-Object Tracking (MOT)~\cite{bewley2016simple, wojke2017simple, fu2025foundation, cai2022memot, meinhardt2022trackformer} is a classical task in computer vision that involves predicting the trajectory of each object in a continuous image sequence while maintaining consistent object identity. Among the various tracking objects, pedestrian tracking has received a lot of attention and has a wide range of applications, such as embodied intelligence, autonomous driving and video surveillance. Most of the existing MOT methods use two paradigms, a tracking-by-detection paradigm~\cite{zhang2022bytetrack, zhang2021fairmot, cao2023observation, aharon2022bot, yang2023hard} based on pretrained detection model and an end-to-end approach~\cite{fu2023denoising, cai2022memot,meinhardt2022trackformer, sun2020transtrack} based on Transformer~\cite{vaswani2017attention}. The former uses existing detection models to perform the detection and subsequently associates the results. The latter retains the hidden states in model as a special tracklet query for each trajectory and directly regresses the location of the tracked object after interacting with the features extracted from the image frames.

In recent years, MOT has made significant progress driven by the fusion of motion and appearance information. Most state-of-the-art methods leverage Kalman filters~\cite{kalman1960new} to model temporal dynamics, updating object motion states (e.g., position, velocity) across frames, while relying on pretrained ReID networks like FastReID~\cite{he2020fastreid} to extract discriminative visual features for appearance matching. However, such dual-cue frameworks face inherent vulnerabilities in challenging environments: motion-based tracking collapses under frequent occlusions (e.g., crowded scenes), where incomplete state observations lead to Kalman filter divergence, while appearance-based association fails in low-quality conditions (e.g., blurry or low-light imagery), as degraded visual features lose discriminative power. These limitations highlight the critical need for robust, context-aware tracking mechanisms that can adapt to diverse real-world scenarios.

% Existing methods rely mainly on the motion or appearance information (or named ReID information) of the object for tracking. For the motion information, many methods use a motion state predictor like Kalman filters~\cite{kalman1960new} to save, update and predict motion information. For the appearance information, methods use a self-designed or off-the-shelf network~\cite{he2020fastreid} to extract the feature of the trajectory. However, both types of information can be non-robust or even completely useless when encountering complex scenarios. For example, when encountering crowded scenarios, the object tends to be frequently occluded, which in turn prevents the motion state matrix from being updated, and thus prevents making accurate predictions. Moreover, when the image becomes blurred, the extracted appearance features of the object will lose their significance.

\begin{figure}[t]
  \centering
  \includegraphics[width=\linewidth]{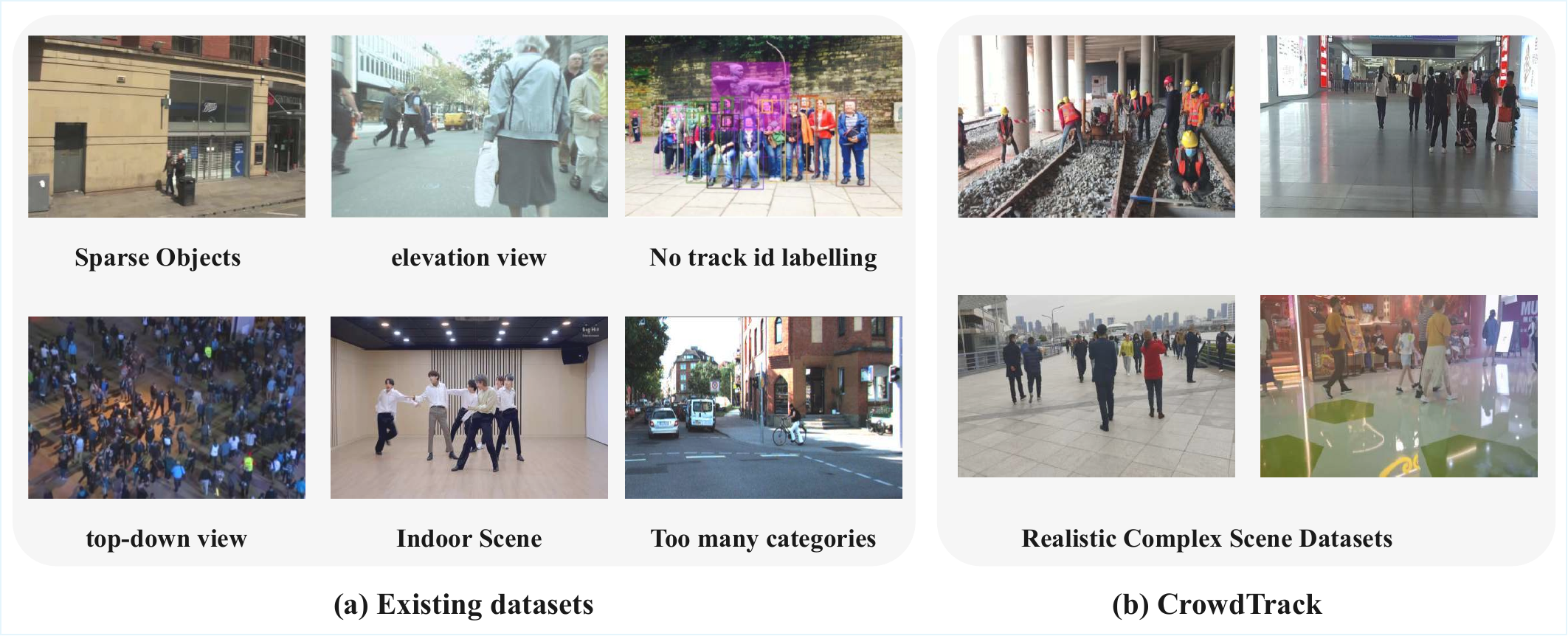}
  \caption{Comparison of our dataset with existing datasets. The images in (a) are from MOT17~\cite{milan2016mot16}, MOT20~\cite{dendorfer2020mot20}, CrowdHuman~\cite{shao2018crowdhuman}, KITTI360~\cite{geiger2012we} and DanceTrack~\cite{sun2022dancetrack}. }
  \label{fig1}
\end{figure}

% Supervised learning using annotated data is still the main model training method in this field, as in other fields. The MOT~\cite{milan2016mot16, dendorfer2020mot20} dataset is the most commonly used and has been released in several versions, SportsMOT~\cite{cui2023sportsmot} and DanceTrack~\cite{sun2022dancetrack} are two recently released large-scale dataset, both of which have become major benchmarks for evaluating MOT methods. However, as Fig.\ref{fig1} shows, each of these datasets has some shortcomings. For the MOT Benchmark, with the development of recent years, the scale has become the biggest drawback of this dataset. While MOT17~\cite{milan2016mot16} is a simpler version, MOT20~\cite{dendorfer2020mot20} has more complex scenes, but most of them are shot from an overhead viewpoint, which leads to the fact that even if the objects are crowded, there are still enough objects to be recognized and detected by the detection model. DanceTrack and SportsMOT, on the other hand, are datasets for specific scenarios, lacking real-life scenarios from everyday life, which can lead to a lack of robustness and generalization of the model. 

Similar to other fields, supervised learning with annotated data remains the dominant model training approach in Multi-Object Tracking (MOT). The MOT dataset~\cite{milan2016mot16, dendorfer2020mot20}, released in multiple versions, along with recent large-scale datasets like SportsMOT~\cite{cui2023sportsmot} and DanceTrack~\cite{sun2022dancetrack}, serve as key benchmarks for evaluating MOT methods. However, as illustrated in Fig.\ref{fig1}, these datasets have notable limitations. The MOT Benchmark's scale has become a major drawback over time; MOT17~\cite{milan2016mot16} is relatively simple, while MOT20~\cite{dendorfer2020mot20} features complex scenes but mainly from overhead views, enabling detection models to identify objects even in crowded scenarios. Meanwhile, SportsMOT and DanceTrack, being specific-scenario datasets, lack everyday real-life contexts, potentially resulting in models with limited robustness and generalization.

% In this paper, we launched CrowdTrack, a large-scale multiple pedestrian tracking dataset. The dataset contains 33 video sequences, about 40,000 image frames and over 700k person annotations. In addition to its scale, our dataset still has many advantages. First, we mainly adopt a real-scene data collection method from a first-person perspective. This fully mimics the data requirements of fields such as robotics and autonomous driving. In this perspective, problems such as crowdedness and occlusion will further affect the performance of the models, bringing new challenges to existing methods. It is worth mentioning that our dataset has both moving lens shots and fixed lens shots, which would give an advantage to methods with camera motion compensation. And all our data comes from real daily life, and the behavior of the object is completely unaffected by us. This more realistic data will be more useful for the practical application of the model. Additional, our dataset contains difficult samples such as occlusion, crowdedness, and even blurring, and the correct labels of these difficult samples will help the model learn how to cope with this complex situations.

In this paper, we present CrowdTrack, a large-scale multi-pedestrian tracking dataset featuring 33 video sequences, around 40,000 image frames, and over 700K person annotations. Beyond its substantial scale, the dataset is designed to address critical gaps in real-world tracking scenarios. It incorporates diverse camera setups, including both moving and fixed lens shots, which can benefit methods incorporating camera motion compensation. All data is collected in unconstrained daily environments, ensuring object behaviors remain natural and unmodified, thus enhancing the dataset’s relevance for practical applications like robotics and autonomous driving. Notably, CrowdTrack includes challenging annotations for complex scenarios such as occlusion, crowding, and blur, providing rich training signals to improve model robustness against real-world complexities.

\begin{figure}[ht]
  \centering
  \includegraphics[width=\linewidth]{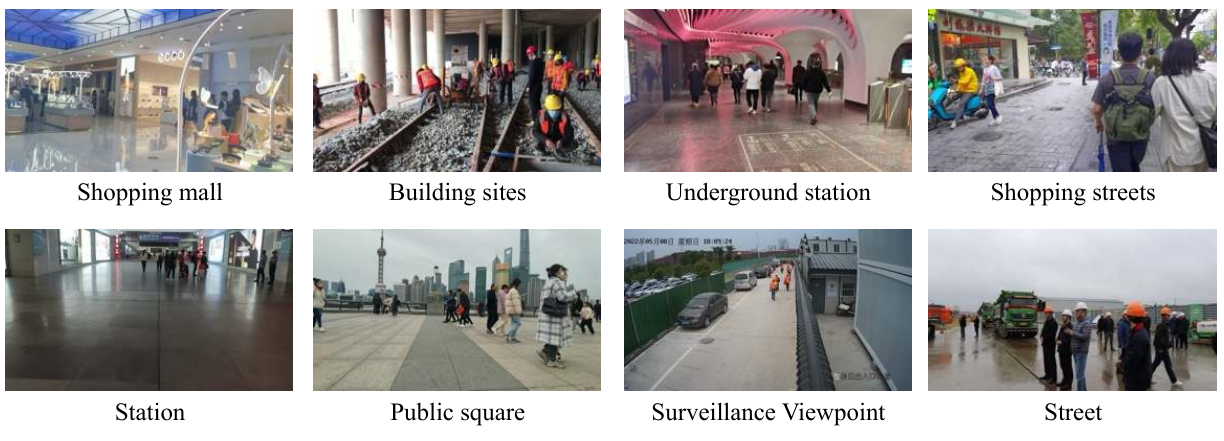}
  \caption{Some sampled scenes from the proposed dataset. All of our data comes from real life, covering a wide range of scenarios, including indoor shopping malls, construction sites, underground stations, outdoor shopping streets, and more. }
  \label{fig2}
\end{figure}

We establish a novel benchmark integrating prevalent multi-object tracking methodologies. Experimental results reveal that the performance of state-of-the-art (SOTA) methods on our dataset experiences varying degrees of degradation compared to their performance on existing benchmarks. This indicates that current SOTA approaches struggle to generalize effectively in complex scenarios characterized by heavy occlusion, motion blur, and dense crowds. Furthermore, we evaluate the capacity of existing foundation models to represent objects within our dataset, offering empirical support for the emerging paradigm of leveraging foundation models to address MOT challenges~\cite{fu2025foundation, li2023ovtrack, li2024matching}.

Our dataset aims to advance MOT research, particularly in enhancing tracking robustness under complex conditions. Simultaneously, we aspire to provide valuable data resources for the development of foundation models with video comprehension capabilities, thereby fostering innovative approaches and investigations into solving video-related tasks using such models. The core contributions of this work are outlined as follows:

\begin{itemize}
    \item We build a new large-scale multi-object tracking dataset, CrowdTrack, which contains a wide range of real-life scenes and includes a variety of difficult samples.
    
    \item We benchmark baseline methods on this newly built dataset with various evaluation metrics, proving that existing methods are still inadequate in solving the multi-object tracking problem in complex scenarios.

    \item  We have comprehensively analyzed our dataset and made attempts to solve difficult scenarios, which helps subsequent research on our dataset.
    
    % Extensive experiments show that our method surpasses state-of-the-art methods in zero-shot settings, and achieves comparable performance in non-zero-shot settings with significantly reduced inference time.
\end{itemize}

\section{Related Work}
\label{sec:relatedwork}
\subsection{Multi-object Tracking Methods}

Tracking-By-Detection (TBD) method consisting of a pretrained detector~\cite{ge2021yolox,ren2015faster} and an assignment strategy. The former detects new objects and the latter assigns them to existing trajectories or initializes them to new trajectories based on positional distance and appearance similarity. The above paradigm was first adopted by SORT~\cite{bewley2016simple}, and Deep SORT~\cite{wojke2017simple} added appearance features as a distance measure. The subsequent methods are optimizing towards more efficient appearance feature extraction~\cite{zhang2021fairmot}, better motion information modeling~\cite{cao2023observation} and more efficient assignment strategies~\cite{yang2023hard}.

Transformer~\cite{vaswani2017attention} has had great success in the field of NLP~\cite{devlin2018bert} and was soon widely used in the field of Computer Vision. Transtrack~\cite{sun2020transtrack} uses the Transformer to replace the components in the TBD paradigm. Trackformer~\cite{meinhardt2022trackformer} uses each active trajectory as a query to do cross-attention with the image frame and regresses directly to the position of the tracking object, while a portion of the query is responsible for detecting new trajectories, as in an object detection network~\cite{zhu2020deformable, carion2020end}, and most of the subsequent work continues the approach. e.g., MeMOT~\cite{cai2022memot} expands the length of the sequence in the memory buffer for each trace, and DNMOT~\cite{fu2023denoising} employs the idea of noising and denoising.

In recent years there have been a number of approaches that differ from both of these paradigms, e.g., OCMOT~\cite{zhao2023object} adopts the idea of object-centric, where each trajectory is treated as a "slot"; DiffusionTrack~\cite{luo2023diffusiontrack} solves the problem using a generative approach based on a diffusion model; and by combining SAM~\cite{kirillov2023segment}, DeAOT~\cite{yang2022decoupling}, and Grounding-DINO~\cite{liu2023grounding}, SAM-Track~\cite{cheng2023segment} implements a multi-object tracking algorithm with multiple interactions. OVTrack~\cite{li2023ovtrack} proposes open-vocabulary MOT, aiming to track all objects in the scene by utilizing CLIP's generalization capabilities for open-world object tracking. MASA~\cite{li2024matching} focuses on fine-grained tracking, exploring instance-level object features using SAM~\cite{kirillov2023segment} and detectors such as Grounding DINO~\cite{liu2024grounding} or YOLOX~\cite{ge2021yolox}. ViPT~\cite{zhu2023visual} explores the effect of adding other modal data to the model's inputs, including heat maps, event information and depth information, and proposes a multimodal model with learnable parameters that account for only 1\% of the total number of parameters.

\subsection{Multi-object Tracking Datasets}

There are many multi-object tracking datasets available, and due to the specificity of pedestrian tracking, there are still some datasets for pedestrian tracking only. The MOT Challenge~\cite{milan2016mot16, dendorfer2020mot20} is the most popular multi-object pedestrian tracking dataset, and has been released in successive MOT15, MOT16, MOT17, MOT20 and other versions. In an effort to raise awareness of the importance of appearance information, DanceTrack~\cite{sun2022dancetrack} has released a series of datasets with dancers which has similar clothing and complex motion. SportsMOT~\cite{cui2023sportsmot}, on the other hand, has published a dataset for sports events. These datasets still have many shortcomings, such as perspective issues, scenario issues, and scale issues. Our dataset, which also focuses on pedestrian tracking, is a large-scale MOT dataset containing a variety of complex real-world scenarios.

There are also many datasets outside of pedestrian tracking that are often used for pre-training and test. The MOTS~\cite{voigtlaender2019mots} and Youtube-VIS~\cite{xu2018youtube} are datasets of Video Instance Segmentation (VIS) task, which require more granular output. KITTI~\cite{geiger2012we}, Waymo~\cite{sun2020scalability} and BDD100K~\cite{yu2018bdd100k} are datasets in the field of autonomous driving, where vehicles are labelled in addition to human. ImageNet-Vid~\cite{deng2009imagenet} and TAO~\cite{dave2020tao} expanded tracking categories to a wider range of categories and OVTrack~\cite{li2023ovtrack} introduced the concept of Open-Vocabulary Multiple Object Tracking, intended to track every object in the video. BenSMOT~\cite{li2025beyond} proposed Semantic MOT benchmark which introduces three extra semantic understanding tasks, Refer-KITTI~\cite{wu2023referring} and Refer-KITTI v2~\cite{zhang2024bootstrapping} proposed Referring MOT task, while CRTrack~\cite{chen2024cross} extends this task to multiple views.

% \subsection{Multi-modal Multi-object Tracking}

% There are many methods that have good performance by training with other modal datasets. In addition to the field of MOT and VIS, there are datasets from other domains that can help the MOT algorithm achieve high performance. CrowdHuman~\cite{shao2018crowdhuman} is a crowded pedestrian detection dataset, which is usually used for the training of object detection models. PoseTrack2018~\cite{andriluka2018posetrack} is a dataset that tracks human joints, and training with it enhances the model's perception of human posture. ViPT~\cite{zhu2023visual} explores the effect of adding other modal data to the model's inputs, including heat maps, event information and depth information, and proposes a multimodal model with learnable parameters that account for only 1\% of the total number of parameters.

\section{CrowdTrack}

\subsection{Dataset Construction}
\textbf{Data collection.} As depicted in Fig. \ref{fig2}, the dataset comprises 33 video sequences collected from real-world environments. To isolate pedestrian dynamics, we prioritized scenes devoid of structural constraints (e.g., pavements) that might influence movement patterns. While typical daily scenarios often involve slow-paced movement and low clothing similarity, we intentionally included footage from building sites to introduce unique challenges: workers’ uniform workwear and helmets suppress facial feature discriminability, thereby emphasizing the importance of gait and body shape features for tracking. All video content undergoes rigorous privacy-preserving processing to obscure identifiable information, ensuring compliance with ethical data-handling standards.

\textbf{Data annotation.} The labeling workflow was executed by commercial partners with professional annotators conducting at least three rounds of quality feedback. Following common dataset conventions, we only annotated 2D-visible human instances: partially occluded individuals were labeled with full-body bounding boxes, while fully occluded persons were excluded. Each tracklet maintains a unique ID throughout its lifecycle, even during temporary disappearances. For human-carried objects, small items (e.g., mobile phones, school bags) are included in person annotations, whereas large carriers (e.g., trolleys) are treated separately, with labeling focused solely on the individual. While we strive to annotate all visible persons, objects below an empirical size threshold are excluded to balance data utility and annotation feasibility.

\subsection{Dataset Statistic}
In this section, we analyze our dataset from multiple dimensions and compare it with existing datasets. Since SportsMOT~\cite{cui2023sportsmot} only labels athletes on the field (contrary to our "all pedestrians labeled" principle), we exclude it from some comparisons to maintain consistency.

\textbf{Scenario Analysis.} As shown in Fig. \ref{fig2}, our dataset encompasses diverse real-life scenarios with pedestrians behaving naturally, including key distinctive scenes: construction sites where workers in uniform clothing, safety helmets, and unconventional movements challenge facial feature-dependent methods; indoor and outdoor shopping malls with frequent human-object occlusions; and additional environments like metro stations, bus stops, and squares to enhance dataset diversity.

\begin{table}[t]
    \centering
    \caption{Scale comparisons between datasets. "Avg." denotes "Average". "Tot." denotes "Total". "T", "B", "I" denote "Tracklets", "Bounding boxes" and "Images", respectively. Since we don't have access to the ground truth of the test set, the total number of objects we derive from the training set.}
    \setlength{\tabcolsep}{4mm}{
    \begin{tabular}{l|cccccc}
    \toprule 
    Dataset & Videos & Avg. T& Tot. T  & Tot. B(k) & Tot. I & FPS  \\
    \midrule
    MOT17~\cite{milan2016mot16} & 14 & 96 & 1342 & 292 & 11235 & 30\\
    MOT20~\cite{dendorfer2020mot20} & 8 & 432 & 3456 &  1652 & 13410 & 25 \\
    DanceTrack~\cite{sun2022dancetrack} & 100 & 9 & 990 &  880 & 105855 & 20 \\
    SportsMOT~\cite{cui2023sportsmot} & 240 & 14 & 3401 & 1629 & 150379&-\\
    Ours & 33 & 158 &5185 & 703 & 34564 & 20\\
    \bottomrule
    \end{tabular}}    
    \label{tab:dataset}
\end{table}

\textbf{Dataset split.} We partition the dataset into training and test sets, following the MOT benchmark’s strategy to evenly distribute similar scenes. The final split comprises 17 training videos and 16 test videos, balanced in video count, tracklets, and pedestrian annotations (see Table \ref{tab:dataset} for details). Compared to MOT17, CrowdTrack significantly surpasses it in scale across all metrics. While MOT20 features complex overhead-view scenes, our dataset introduces crowded first-person-view scenarios, though with a smaller average object count. In contrast to DanceTrack, CrowdTrack contains more densely packed scenes and achieves comparable object numbers with approximately three times fewer images, highlighting its efficiency in capturing challenging tracking dynamics.

\textbf{Object Motion Analysis.} Motivated by DanceTrack~\cite{sun2022dancetrack}, we analyzed the motion information in the dataset by three metrics and compared our dataset with other MOT datasets.  

First, we compute the \textbf{IOU Scores} of the object's bounding box in two adjacent frames. The IOU score is calculated as follows:

\begin{equation}
  S_{iou} = \frac{1}{N(T-1)}\sum_i^N\sum_{t=1}^{T-1}\mathtt{IoU}(\mathbf{B}_i^t,\mathbf{B}_i^{t+1})
  \label{siou}
\end{equation}
where $N$ denotes the number of objects in the sequence, $T$ denotes the length of the sequence, and the bounding box of object $i$ at frame $t$ we denote as $\mathbf{B}_i^t$. $\mathtt{IoU}(\cdot,\cdot)$ denotes the IoU score. $S_{iou}$ is related to the frame rate of the video and the motion pattern of the object itself, the higher the score, the slower the object moves, which in turn reflects a simpler motion pattern. 

\begin{figure*}[t]
     \centering
     \begin{subfigure}[b]{0.45\textwidth}
         \centering
         \includegraphics[width=\textwidth]{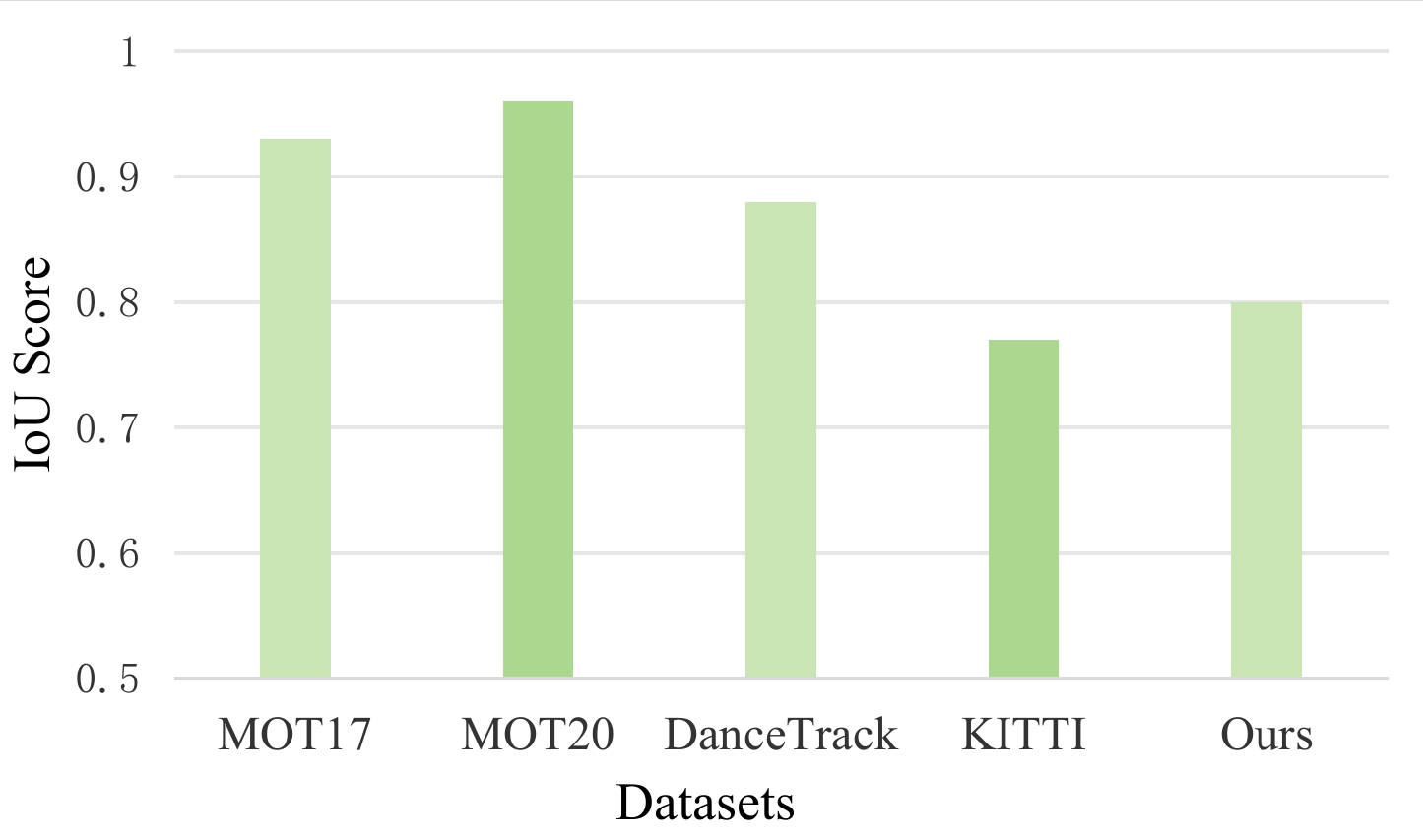}
         \caption{IoU between consecutive frames}
         \label{iou}
     \end{subfigure}
     \hfill
     \begin{subfigure}[b]{0.45\textwidth}
         \centering
         \includegraphics[width=\textwidth]{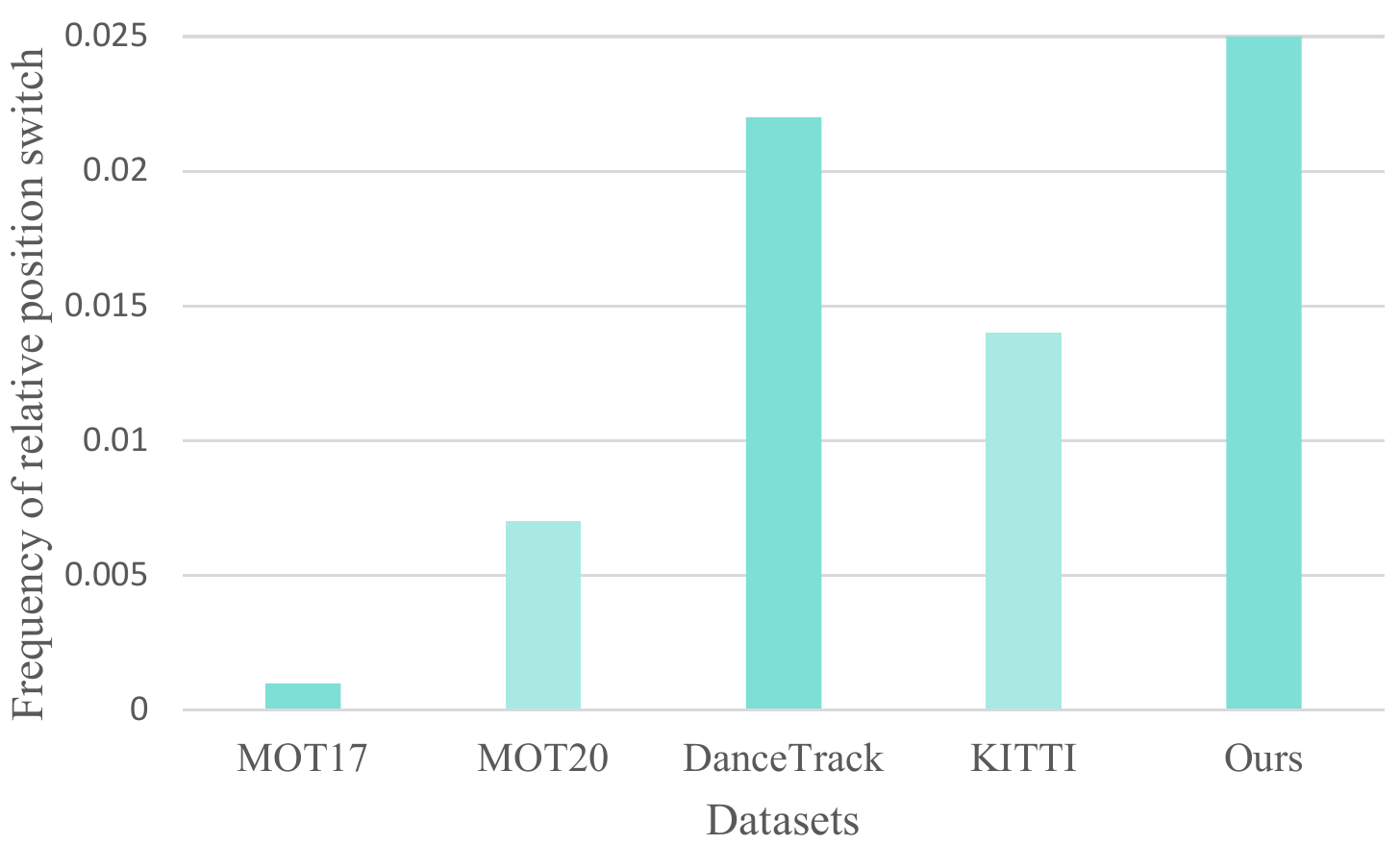}
         \caption{Frequency of relative position switching}
         \label{switch}
     \end{subfigure}
     \hfill
     \begin{subfigure}[b]{0.45\textwidth}
         \centering
         \includegraphics[width=\textwidth]{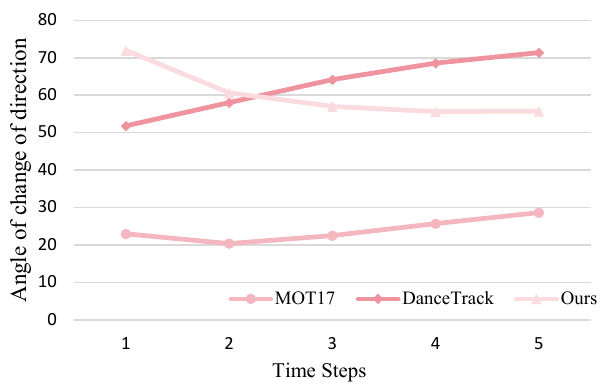}
         \caption{Magnitude of change in direction}
         \label{angle}
     \end{subfigure}
     \hfill
     \begin{subfigure}[b]{0.45\textwidth}
         \centering
         \includegraphics[width=\textwidth]{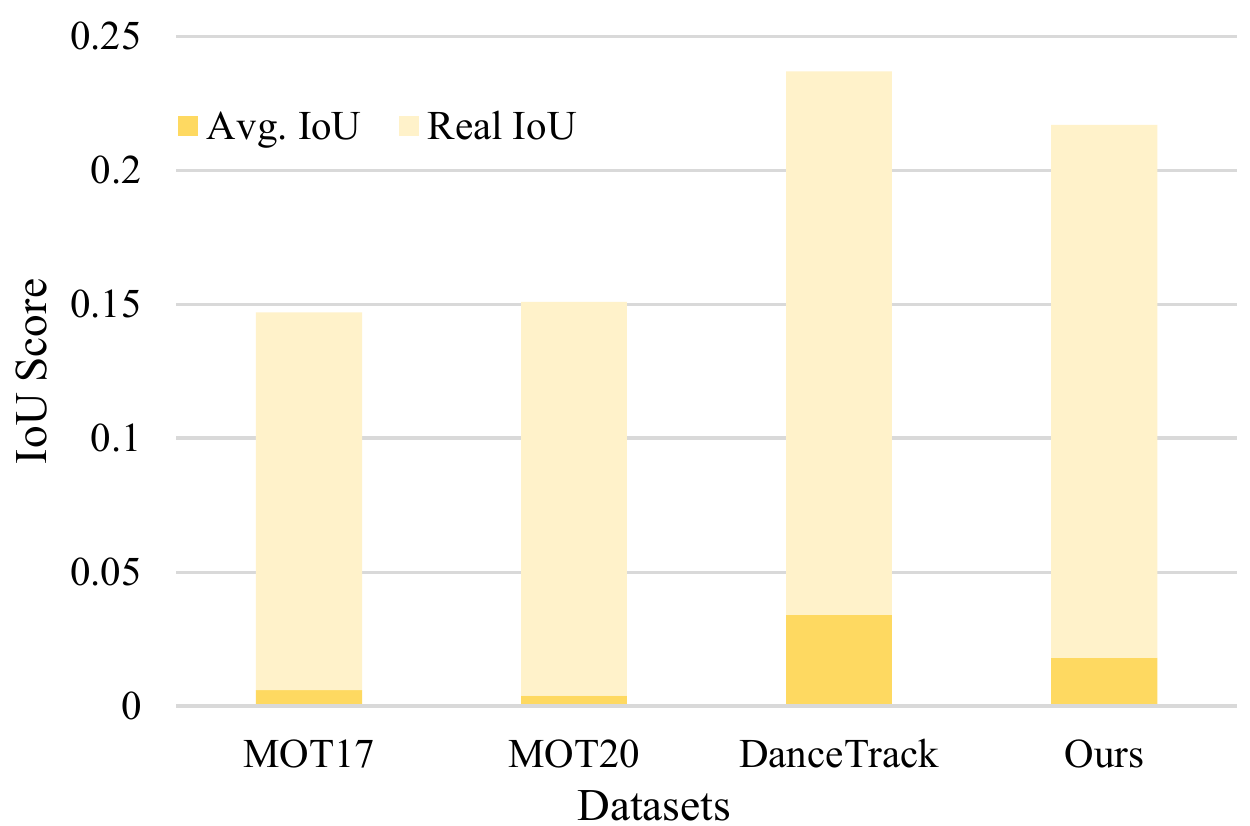}
         \caption{Crowdedness analysis}
         \label{crowd}
     \end{subfigure}
     
        \caption{Some quantitative analyses of the proposed dataset. (a) compares the average overlap of object positions between adjacent frames. (b) compares the average frequency of relative position switching. (c) compares the average angular change in the direction of the objects between units of time and (d) compares the average crowdedness and true crowdedness.}
        \label{fig:vis}
\end{figure*}

We then measure the relevance of the objects to each other, and like in DanceTrack, we use \textbf{Frequency of Relative Position Switch} to represent the metric. This metric measures the number of times the object switches relative to the position of another object in two adjacent frames, regardless of whether the switch occurs around the x-axis or the y-axis position, which is counted as only one time. Specifically, the score can be obtained by the following formula:

\begin{equation}
  S_{sw} = \frac{\sum_i^N\sum_{j\neq i}^N\sum_{t=1}^{T-1}\mathtt{sw}(\mathbf{B}_i^t,\mathbf{B}_j^t,\mathbf{B}_i^{t+1},\mathbf{B}_j^{t+1})}{2N(T-1)(N-1)}
  \label{ssw}
\end{equation}
where $\mathtt{sw}(\cdot)$ is an indicator function, where $\mathtt{sw}(\cdot)=1$ if the two objects swap their left-right relative position or top-down relative position on the adjacent frames. This metric statistics the complexity of object movement in the dataset. The more frequent the switching, the more irregular the movement of the object is, and the more challenging it is for the model to model the motion of the object.

Finally, we analyze the motion complexity of the objects. We measure the complexity of the motion of the object by the \textbf{Direction Switching Angles}. This indicator measures the magnitude of change in the direction of an object over successive time periods. This indicator can be calculated as:

\begin{equation}
  S_{angle} = \frac{1}{N(T-2\tau)}\sum_i^N\sum_{t=1}^{T-2\tau} \mathtt{arccos}{(\mathbb{N}(\overrightarrow{\mathbf{B}_i^t\mathbf{B}_i^{t+\tau}}),\mathbb{N}(\overrightarrow{\mathbf{B}_i^{t+\tau}\mathbf{B}_i^{t+2\tau}}))} 
  \label{ssw}
\end{equation}
where $\mathbb{N}(\cdot)$ denotes vector normalization function, $\tau$ denotes time span and $\overrightarrow{\mathbf{B}_i^t\mathbf{B}_i^{t+\tau}}$ denotes the vector formed by the center points of the bounding box of the object $i$ at frames $t$ and $t+\tau$. This metric calculates the motion irregularities of individual objects, and to explore this over a longer period, we have taken different values for $\tau$. The greater the angle at which the object switches direction per unit of time, the motion of the object is more irregular.

We compared the datasets based on these three metrics and placed the results in Fig.\ref{iou} - Fig.\ref{angle}. For the IoU of consecutive frames, our dataset is ahead of MOT17, MOT20 and DanceTrack. There is a strong correlation between this metric and the frame rate. But even compared to DanceTrack at the same frame rate, our dataset still has a lower IoU score. Since the KITTI dataset focuses on autonomous driving, the camera will also move faster, resulting in a faster relative movement of the object, so it is reasonable that we are not as good as KITTI in this metric. Our dataset has the largest relative movement frequency of all compared datasets, demonstrating the complexity of the dataset scenario. And interestingly, our dataset produces completely different results on the third metric than MOT17 and DanceTrack. As the time step increases, the objects in the dataset show a slight change in direction (e.g., if you're shopping in a mall, you may be moving in a general direction but you're constantly attracted to things on either side and change direction briefly). This further encourages the model to model the motion of the object from a longer temporal magnitude. When moving along a more regular non-straight line, the angular change over long periods of time is greater than over short periods of time, which explains the trend in the remaining two datasets.

\textbf{Crowdedness analysis.} We use the average IoU between objects in the same frame to evaluate how crowded a dataset is. In order to avoid the case of too many zeros in the IoU between objects, we also calculated only the average IoU between objects that overlap each other, denoted as \textbf{Real IoU}. The results are shown in Fig. \ref{crowd}. The metric is calculated as follows:

\begin{figure}[t]
  \centering
  \includegraphics[width=\linewidth]{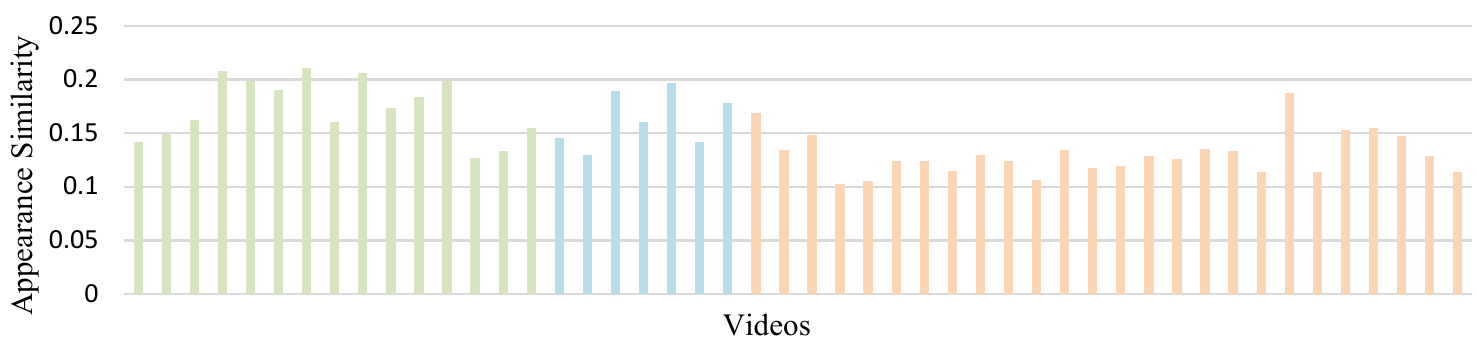}
  \caption{Average similarity of validation set sequences in each dataset.The green, blue and orange colors represent our datasets, MOT17 and DanceTrack respectively. The mean similarity for the three datasets is 0.18, 0.17, and 0.13, respectively.}
  \label{figreid}
\end{figure}

\begin{equation}
  S_{RIoU} = \frac{1}{T}\sum_{t=1}^T\frac{1}{N_t^2}\sum_i^{N_t}\sum_{j\neq i}^{N_t}\mathtt{IoU}(\mathbf{B}_i^t,\mathbf{B}_j^t)),  \quad \mathtt{IoU}(\mathbf{B}_i^t,\mathbf{B}_j^t)>0
  \label{riou}
\end{equation}
Where $N_t$ is the number of objects in frame $t$. While MOT20 exhibits the highest average object density, its overhead viewing angle limits object overlap, resulting in the lowest average IoU despite dense annotations. In contrast, DanceTrack’s sequences achieve high Real IoU due to stage constraints that position actors in predefined locations for performance effects. Notably, our dataset—designed to align with the MOT series’ general framework—matches DanceTrack’s Real IoU performance while introducing more naturalistic crowd dynamics and first-person perspective challenges.

\textbf{Object appearance analysis.} We use a pre-trained Vision Transformer~\cite{dosovitskiy2020image} to extract the appearance feature vector of the objects in existing datasets and compare the similarity between the objects in the same frame. Specifically, we use the following formula to calculate the similarity:
\begin{equation}
  S_{App} = \frac{1}{T}\sum_{t=1}^T\frac{1}{N_t^2}\sum_i^{N_t}\sum_{j\neq i}^{N_t}1-\mathtt{Norm}(\mathtt{App}(\mathbf{B}_i^t))\cdot \mathtt{Norm}(\mathtt{App}(\mathbf{B}_j^t))
  \label{riou}
\end{equation}
Where $\mathtt{App}(\cdot)$ is ReID model. The result for MOT17, DanceTrack and ours are shown in Fig.\ref{figreid}. Objects in DanceTrack achieve the highest appearance similarity due to similar costumes and dance moves. Because both our dataset and the MOT dataset are derived from real-world scenes, they exhibit similar appearance similarity.

\subsection{Evaluation Metrics}
We employ a multivariate approach to evaluate MOT methods, starting with individual metrics such as Mostly Tracked (MT), Mostly Lost (ML), False Negative (FN), False Positive (FP), and ID Switch (ID.Sw). These single metrics are then used to calculate composite metrics, with MOTA~\cite{bernardin2008evaluating} historically serving as the dominant metric. However, due to the inconsistent order of magnitude among FP, FN, and ID.Sw in results, MOTA often overemphasizes detection performance while underweighting algorithmic assignment strategies. In recent years, HOTA~\cite{luiten2021hota} has emerged as a critical evaluation metric for MOT, even becoming the primary benchmark for datasets like DanceTrack and BDD100K. Accordingly, we adopt HOTA as the core metric for our benchmark to enable a more comprehensive assessment of MOT methods.

\section{Experiments}
% In this section, we evaluate the existing methods in our dataset. In addition to this, we explored on the dataset how to perform multi-object tracking in complex scenes.

\subsection{Settings}
We mainly compare the performance of the methods on MOT17~\cite{milan2016mot16}, DanceTrack~\cite{sun2022dancetrack} and on our dataset. For our dataset and MOT17, since there is no validation set, we followed the practice in methods such as CenterTrack~\cite{zhou2020tracking} and FairMOT~\cite{zhang2021fairmot} by dividing half of the training set as the validation set. As for DanceTrack, we use the division constructed by the paper.
We utilized PyTorch to develop our experiments and carried out them on 8 A100 GPUs. When it comes to reproducing and testing existing methods, we use the default configurations of the methods and ensure that they are identical across the evaluation of the datasets.

\subsection{Benchmark Results}

\begin{table}[t]
    \centering
    \caption{Comparison of the performance of existing methods on a test set of each dataset. The best performance is highlighted in bold. Indicators not indicated in the original paper are denoted by "$-$". 
}
    \scalebox{1}{
    \setlength{\tabcolsep}{1mm}
    \begin{tabular}{l|cccc|ccc|ccc}
    \toprule 
    \multirow{2}{*}{Methods} & &\multicolumn{3}{c|}{MOT17} & \multicolumn{3}{c|}{DanceTrack} & \multicolumn{3}{c}{Ours} \\
    \cmidrule{3-11}
     & &MOTA & HOTA & ID.Sw & MOTA & HOTA & IDF1 & MOTA & HOTA & ID.Sw \\
    \midrule
    FairMOT~\cite{zhang2021fairmot} & & 73.7 & 59.3 & 3303 & 82.2 & 39.7 & 40.8 & 34.66 & 39.52 & \textbf{1857} \\
    TraDeS~\cite{wu2021track} & & 69.1 & 52.7 & 3555 & 86.2 & 43.3 & 41.2 & 44.14 & 37.08 & 3431 \\
    MeMOTR~\cite{gao2023memotr} & & 72.8 & 58.8 & - & 85.4 & \textbf{63.4} & \textbf{65.5} & 52.97 & 46.61 & 1937 \\
    DiffusionTrack~\cite{luo2023diffusiontrack} & & 77.9 & 60.8 & 3819 & 89.3 & 52.4 & 47.5 & 58.96 & 45.51 & 6660 \\
    Trackformer~\cite{meinhardt2022trackformer} & & 74.1 & 57.3 & \textbf{1532} & - & - & - & 62.04 & 47.86 & 3218 \\
    OC-SORT~\cite{cao2023observation} & & 78.0 & \textbf{63.2} & 1950 & 89.6 & 54.6 & 54.6 & 62.29 & 52.79 & 1879 \\
    ByteTrack~\cite{zhang2022bytetrack} & & \textbf{80.3} & 63.1 & 2196 & \textbf{89.5} & 47.3 & 53.9 & \textbf{66.69} & \textbf{53.66} & 2531 \\

    \bottomrule
    \end{tabular}
    }
    
    \label{tab:benchmark}
\end{table}

We evaluated existing state-of-the-art (SOTA) methods on our dataset and compared their performance against results on other benchmarks. The tested methods span diverse paradigms, including Tracking-By-Detection (TBD) approaches, Transformer-based models, and Generative Model frameworks (e.g., diffusion models~\cite{yang2023diffusion}). Detailed results are presented in Table \ref{tab:benchmark}.

\begin{wraptable}{tr}{0.55\textwidth} % r表示右侧，0.5\textwidth为表格宽度
    \centering
    \caption{Effect of different information on final MOT accuracy.}
    \begin{tabular}{l|cccc}
    \toprule 
    Metric & MOTA & HOTA & IDF1 & ID.Sw \\
    \midrule
     IoU & 91.6 &\textbf{73.3} & 67.2 & 7125\\
     Motion& \textbf{91.9} & 70.3 & \textbf{71.4} & \textbf{1653}\\
     Appearance& 86.2 & 35.8 & 28.2 & 25725 \\
    \bottomrule
    \end{tabular}
    
    \label{tab:distance}
\end{wraptable}

As shown in the table, first, all methods exhibit varying degrees of performance degradation across all metrics compared to other datasets, highlighting that the complex scenarios in our dataset pose new challenges to current MOT algorithms. Second, MOTA declines more significantly than HOTA, indicating that object detection is more challenging in our dataset and underscoring the need to improve detection methods for small and occluded objects—not just association-phase techniques. Notably, while DiffusionTrack achieves comparable results on other datasets, its accuracy drops substantially on ours, particularly in the ID Switch metric, which is 2–4× higher than other methods. This suggests that leveraging novel architectures (e.g., Mamba~\cite{gu2023mamba}, diffusion models, Slot Attention~\cite{locatello2020object}) to address MOT remains a promising research direction.

% \begin{table*}[t]
% \begin{floatrow}
% \capbtabbox{
% \begin{tabular}{l|cccc}
%     \toprule 
%     Metric & MOTA & HOTA & IDF1 & ID.Sw \\
%     \midrule
%      IoU & 91.6 &\textbf{73.3} & 67.2 & 7125\\
%      Motion& \textbf{91.9} & 70.3 & \textbf{71.4} & \textbf{1653}\\
%      Appearance& 86.2 & 35.8 & 28.2 & 25725 \\
%     \bottomrule
%     \end{tabular}
% }{
%  \caption{Effect of different information on final MOT accuracy.}
%  \label{tab:distance}
% }
% \capbtabbox{
% \begin{tabular}{l|ccc}
% \toprule
% Strategy & MOTA & HOTA & ID.Sw \\
% \midrule
% Baseline~\cite{zhang2022bytetrack}&66.69 & 53.66 & \textbf{2531}\\
% BBox Reuse & \textbf{68.32} & \textbf{54.15} &  2840\\
% \bottomrule
% \end{tabular}
% }{
%  \caption{Performance with bounding box reuse strategy.}
%  \label{occ}
%  \small
% }
% \end{floatrow}
% \end{table*}

\subsection{Distance Metric}

We tested the results of using different types of information on our dataset, including modeling with motion and appearance information and using only IoU information, with ByteTrack~\cite{zhang2022bytetrack} as the baseline to ensure consistent association strategies and ground truth as detection results to eliminate interference from other factors. As shown in Table \ref{tab:distance}, using IoU alone as a distance metric achieved the highest HOTA, likely because it directly captures spatial overlap in dense scenarios, while motion information performed optimally in other metrics but was less effective in HOTA due to the use of corrected (rather than ground truth) positions for final output evaluation, which limited the accuracy of motion-based adjustments. Additionally, the frequent occlusions in our dataset caused appearance features alone to perform poorly, as visual discriminability was degraded under such conditions.

\subsection{Discussion}
% In this section, we discuss some future works that may be promising on our dataset or the Multi-object Tracking field.

\begin{figure*}[t]
     \centering
     \begin{subfigure}[b]{0.5\textwidth}
         \centering
         \includegraphics[width=\textwidth]{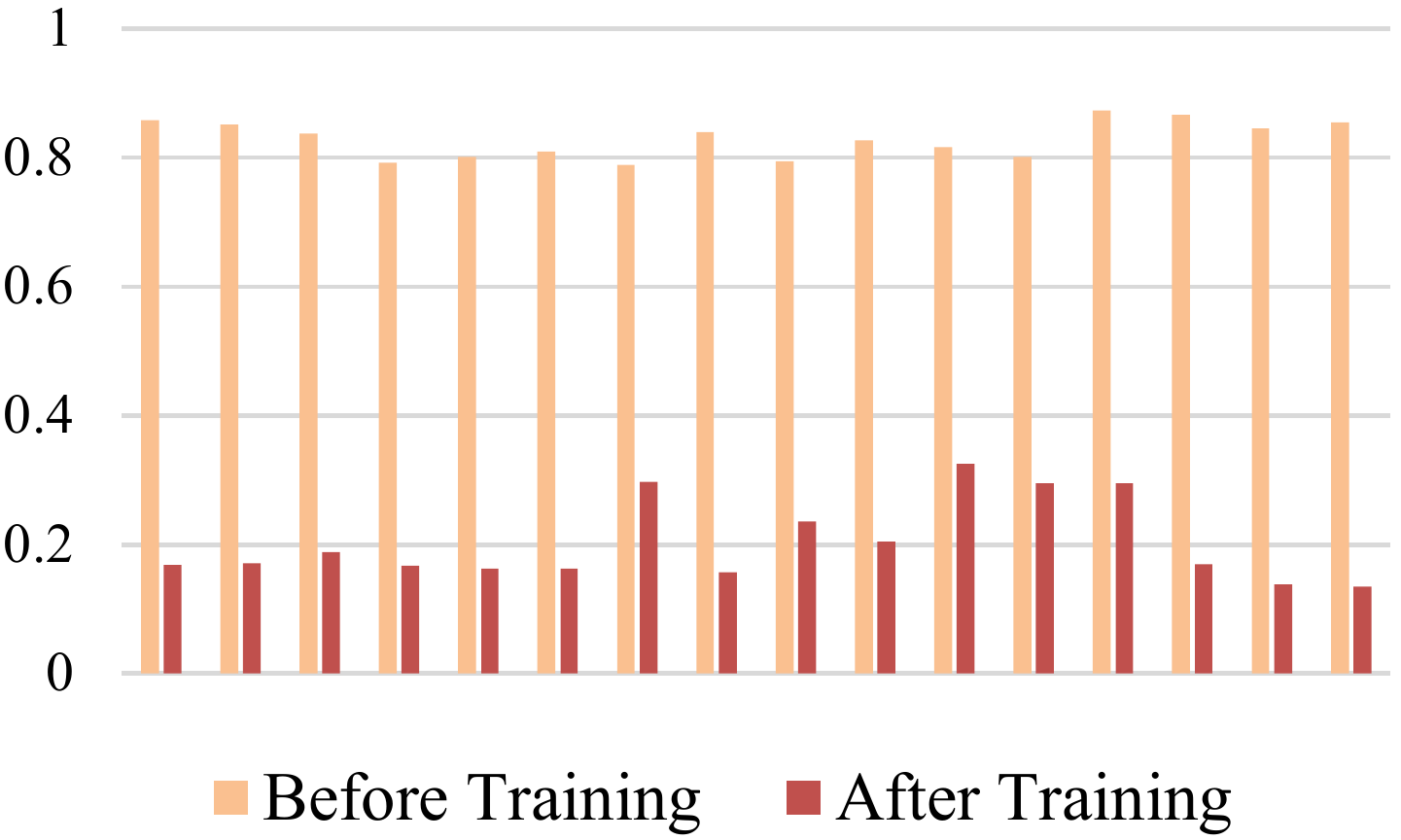}
         \caption{Appearance similarity}
         \label{gap}
     \end{subfigure}
     \hfill
     \begin{subfigure}[b]{0.35\textwidth}
         \centering
         \includegraphics[width=\textwidth]{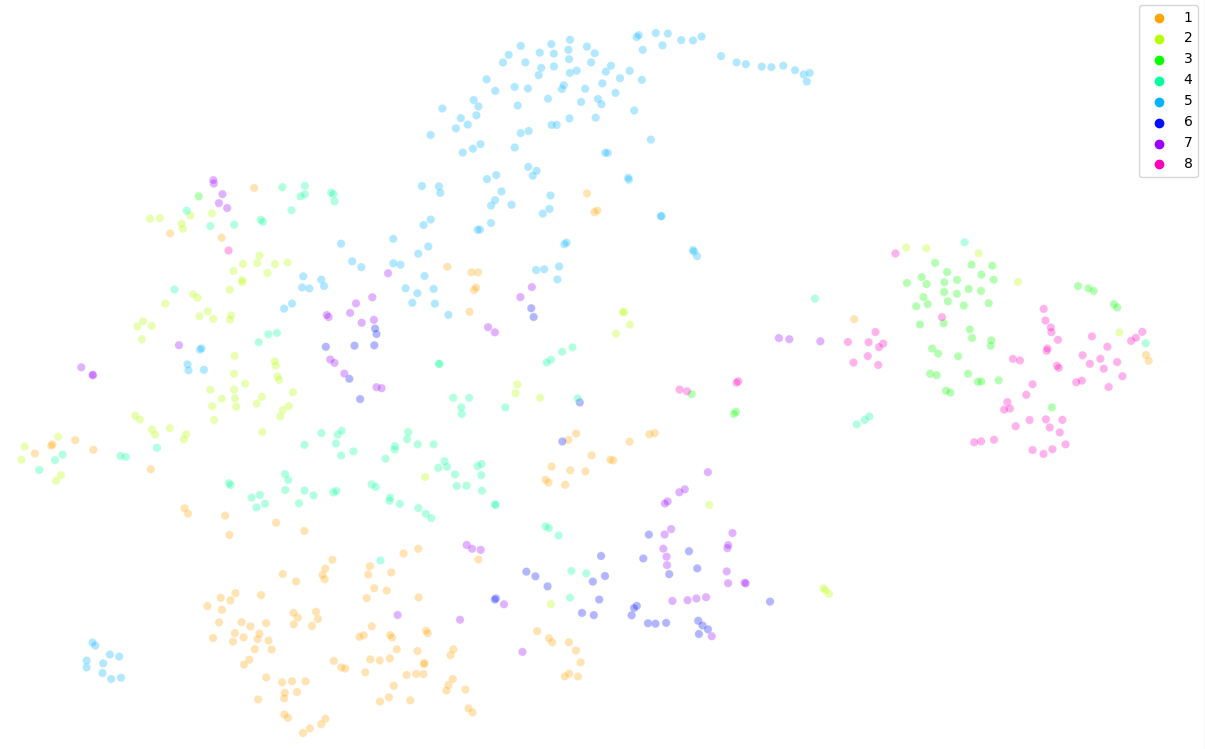}
         \caption{Features on our dataset.}
         \label{tsne}
     \end{subfigure}
     \hfill
     \begin{subfigure}[b]{0.6\textwidth}
         \centering
         \includegraphics[width=\textwidth]{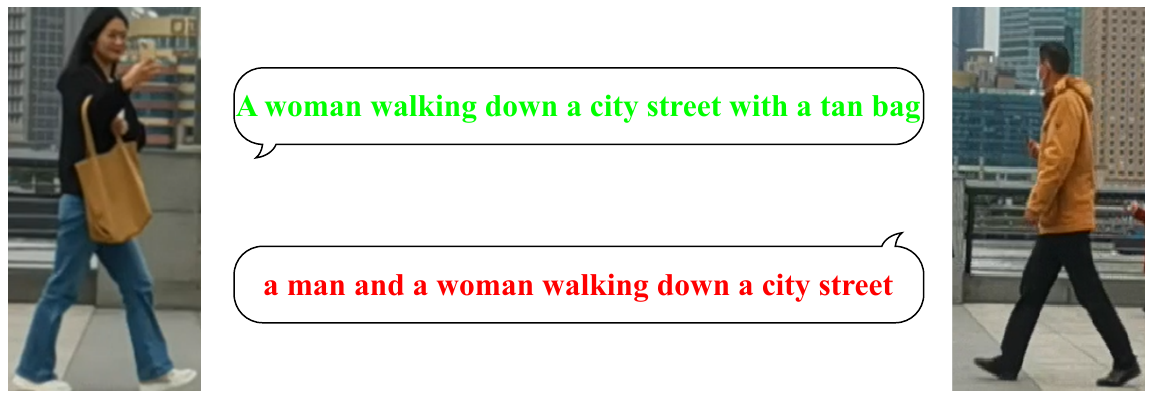}
         \caption{Two captions from BLIP2.}
         \label{ca}
     \end{subfigure}
     \hfill
     \begin{subfigure}[b]{0.35\textwidth}
         \centering
         \includegraphics[width=\textwidth]{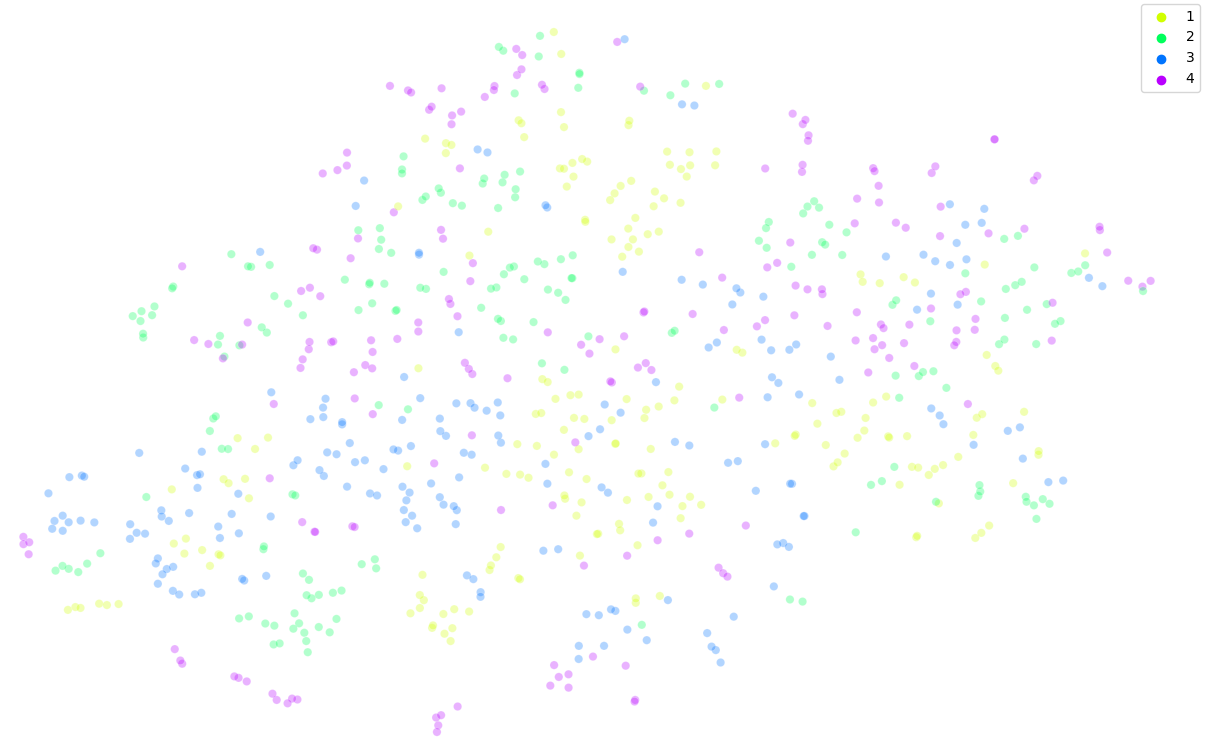}
         \caption{Features on DanceTrack.}
         \label{dan}
     \end{subfigure}
     
        \caption{Some visualization from foundation models on our datasets. (a) shows the similarity comparison before and after training. (b) and (d) shows the appearance features visualization using t-SNE. (c) shows a successful and a failed caption case by the BLIP2~\cite{li2023blip}.}
        \label{fig:vis}
\end{figure*}

\textbf{What can foundation models do with our datasets?} 
% First of all, our dataset can provide weak support for training large foundation models that require huge amounts of data. We also hope that the proposed dataset can contribute to the progress of large foundation models in video understanding and as unsupervised dataset in other fields.
Our dataset can be used for tasks like visual grounding, captioning, and appearance feature extraction. For example, using the BLIP2~\cite{li2023blip} model for captioning random characters shows it can notice individual details (e.g., identifying a girl’s tan bag) but may make errors (e.g., misrecognizing one man as two in an image, see Fig. \ref{ca}). For more Caption results of VLM on this dataset, please refer to the supplementary materials.

We then attempt to extract the appearance features of the objects using a large model. Specifically, we employ CLIP's~\cite{radford2021learning} frozen image encoder as the backbone and append two trainable linear layers behind it to construct a lightweight ReID network architecture. we only train the two linear layers, preserving multi-modal semantic alignment capabilities while reducing computational complexity.
To quantitatively evaluate the impact of training on feature representation, we compared the similarity distribution gaps between the frozen image encoder and the end-to-end trained model, with results presented in Fig.\ref{gap}. Experimental data shows that the trained feature space significantly improves inter-class separability, verifying the effectiveness of trainable layers for modality adaptation. Meanwhile, using t-SNE~\cite{van2008visualizing}, we visualized the appearance features of the trained model for several objects across 100 consecutive frames in Fig.\ref{tsne} and \ref{dan}. The visualizations reveal that although training enhances the model's ability to distinguish different objects, the final feature distributions still exhibit significant intra-class confusion due to the dataset containing only pedestrians.
These results highlight the limitation of the current approach: when object categories exhibit high visual similarity, relying solely on pretrained visual features struggles to achieve robust appearance modeling. 
% Therefore, how to effectively integrate the semantic prior knowledge of large multi-modal models to build more discriminative feature extraction networks remains a critical future research direction.

\subsection{Limitation}
Several future work directions exist. First, our dataset lacks multi-modal annotations (e.g., text, pose, segmentation) compared to datasets like COCO. Adding such annotations could enable richer feature learning. Second, a more robust, dataset-specific model is needed to address its challenges (e.g., high visual similarity). Lastly, expanding the dataset’s scale and diversity (e.g., varied appearances, scenarios) will improve model generalizability. These are key areas for future research.

\section{Conclusion}

We introduces CrowdTrack, a novel large-scale multiple pedestrian tracking dataset that obtained from real-world scenarios. With its extensive scale and complexity, CrowdTrack presents formidable challenges to existing multi-object tracking algorithms, especially in handling dense crowds and occlusions. We conduct in-depth analysis of the dataset and rigorously test state-of-the-art methods, revealing notable performance shortfalls in extreme conditions. Additionally, we explore future research directions for multi-object tracking. Our goal is for CrowdTrack to serve as a pivotal benchmark for developing advanced algorithms in challenging scenarios, thereby driving the progress of multimodal foundation models in video understanding.
\bibliographystyle{unsrtnat}
\bibliography{neurips_2025}

\end{document}